\documentclass{article}

\usepackage{subcaption}

\usepackage[utf8]{inputenc} 
\usepackage[T1]{fontenc}    
\usepackage{hyperref}       
\usepackage{url}            
\usepackage{booktabs}       
\usepackage{amsfonts}       
\usepackage{nicefrac}       
\usepackage{microtype}      

\usepackage{graphicx}
\usepackage{xcolor}
\PassOptionsToPackage{warn}{textcomp}
\usepackage{textcomp}

\usepackage[nonatbib,preprint]{neurips_2020}

\usepackage{scalerel,stackengine,amsmath}
\newcommand\equalhat{\mathrel{\stackon[1.5pt]{=}{\stretchto{%
    \scalerel*[\widthof{=}]{\wedge}{\rule{1ex}{3ex}}}{0.5ex}}}}
\usepackage{bm}

\title{Measuring Information Transfer in Neural Networks}

\author{%
  Xiao Zhang \\
  Department of Electronics Engineering\\
  Tsinghua University\\
  \texttt{xzhang19@mails.tsinghua.edu.cn} \\
  \And
  Xingjian Li \\
  Baidu Research\\
  \texttt{lixingjian@baidu.com} \\
  \And
  Dejing Dou \\
  Baidu Research\\
  \texttt{doudejing@baidu.com} \\
  \And
  Ji Wu \\
  Department of Electronic Engineering\\
  Tsinghua University\\
  \texttt{wuji\_ee@mail.tsinghua.edu.cn} \\
}

\begin{document}

\maketitle

\begin{abstract}
Quantifying the information content in a neural network model is essentially estimating the model's Kolmogorov complexity. Recent success of prequential coding on neural networks points to a promising path of deriving an efficient description length of a model. We propose a practical measure of the generalizable information in a neural network model based on prequential coding, which we term \textit{Information Transfer} ($L_{IT}$). 
 Theoretically, $L_{IT}$ is an estimation of the generalizable part of a model's information content. In experiments, we show that $L_{IT}$ is consistently correlated with generalizable information and can be used as a measure of patterns or ``knowledge'' in a model or a dataset. Consequently, $L_{IT}$ can serve as a useful analysis tool in deep learning. In this paper, we apply $L_{IT}$ to compare and dissect information in datasets, evaluate representation models in transfer learning, and analyze catastrophic forgetting and continual learning algorithms. $L_{IT}$ provides an information perspective which helps us discover new insights into neural network learning.

\end{abstract}

\section{Introduction}

In machine learning, learning a model can be understood as a process of gaining information about the model parameters from the data \cite{mackay2003information}. The learned model is able to generalize because the information gained captures regularities in the data. 
For deep learning models, the information, or ``knowledge'' learned from a large amount of data is crucial for its effectiveness in a wide range of tasks, for example in computer vision \cite{huh2016makes} and NLP \cite{devlin-etal-2019-bert}. Life-long machine learning \cite{chen2018lifelong} even identifies the ability to accumulate and reuse information an indispensable aspect of AI. A natural question to ask is how to measure the information content in a model, or the amount of information transferred from a certain dataset to the model during the process of training.

The pioneering work of Minimum Message Length (MML) \cite{wallace2005statistical} and Minimum Description Length (MDL) \cite{grunwald2007minimum} view learning as data compression, and they give a framework and guidelines on finding the optimal codelength of model and data. In algorithmic information theory, the minimum codelength (or description length) can be used as a measure of the amount of information.

For measuring the description length of a neural network model together with data, it has been shown that prequential codes in practice can achieve efficient codelength, sometimes much better than variational or Bayesian codes \cite{NIPS2018_7490, DBLP:journals/corr/abs-2003-12298}. It is also showed \cite{yogatama2019learning} that the prequential codelength of a target dataset is linked to the generalizability of a transfer learning model. However, as the prequential code is a one-part code \cite{grunwald2007minimum} which encodes both the model and the data, one cannot directly derive the codelength of the model alone.

Although the codelength of a model can be directly given by a two-part code, current implementations of two-part codes for neural networks cannot give practical efficient codes \cite{graves2011practical}. We take an approach based on prequential codes which are proved to work well with neural networks. Instead of finding a theoretical bound of a model's Kolmogorov complexity, in this paper we explore finding a \textit{useful} approximation of information content of a neural network model. Although the approximation can be derived from Kolmogorov complexity, because of a lack of analysis on the convergence properties of prequential coding, the quality of the approximation is hard to guarantee theoretically. 

Therefore, we also include empirical study to show the validity of this approximation, and whether it can serve as a measure of the generalizable information in a model.
%
As we will show in this paper, this approximation is useful in many scenarios for the analysis and understanding of the learning process, from the perspective of information.


The main contributions of this paper are summarized as follows:
\begin{itemize}
	\item We propose a practical measure that approximates the amount of generalizable information in a neural network model, which we term \textit{Information Transfer}. We improved over existing methods by specifically estimating the \textit{generalizable} information in a model, which is usually the central concern in learning.
	\item Empirically, we demonstrate the ability of \textit{Information Transfer} to measure the information gained in a neural network model during training. We can also use it to measure the information content of a task or a dataset. This allow us to analyze the information structure of a dataset and the dynamics of learning.
	\item We perform analysis of transfer learning and continual learning from an informational perspective using \textit{Information Transfer}, which brings new insight into the nature of transfer learning and continual learning, such as how information is transferred and forgotten.
\end{itemize}

\section{The Information Transfer Measure}
Information content can be measured using the length of the shortest code that can reproduce the data, which is also known as the Kolmogorov Complexity \cite{li2008introduction} of the data. To compress the data, one could learn a model to predict the data, and only encode the residues of prediction. Using prequential codes (or online codes) is a model-agnostic way to encode data together with a model.

The basic idea of prequential coding is illustrated in the following example. Imagine Alice wants to send the labels $y_{1:n}$ of a dataset $D$ to Bob, and they both agree on an initial model $\theta_0$. Each time Alice sends one example, she first uses her model to predict the label $y_i$ and encodes $y_i$ using the output distribution of $\theta_{i-1}$. Then she optimizes and updates her model to $\theta_i$ with example $i$. Bob also uses his model to make predictions about $y_i$ and then recover the true $y_i$ with the help of the code he receives. Afterward he updates his model in the same fashion. We denote the length of such a code as $L^{\text{preq}}_{\theta_0}$:
\begin{align}
	L^{\text{preq}}_{\theta_0}(y_{1:n}|x_{1:n}) :=&\ - \sum_{i=1}^n\log p_{\theta}(y_i|x_{1:i},y_{1:i-1})\\
	\label{equ:preq}
	=& - \sum_{i=1}^n\log p_{\theta_{i-1}}(y_i|x_i)
\end{align}
where $\theta_i$ is the parameter of the model learned on samples $\{x_{1:i},y_{1:i}\}$. After Bob receives the whole dataset, he also ends up with the same model that Alice has. In this sense, the prequential code encodes the model together with the data without explicitly encoding the model.

If we use $M$ to denote model $\theta_n$ and use $K$ for Kolmogorov complexity, then $L^{\text{preq}}(y_{1:n}|x_{1:n})$ can be seen as an approximation of $K(D)$ or $K(M,D)$. The latter two equal because the conditional Kolmogorov complexity $K(M|D)=0$, if we assume $\theta_0$ and the training procedure of $M$ is known.


%

\subsection{From $K(M)$ to $L_{IT}$}
\label{sec:lit}
In learning problems we are concerned with how much information is learned by the model. A straight-forward appraoch is to measure $K(M)$:
\begin{align}
	K(M) &= K(M,D)-K(D|M)\\
	&\approx L^{\text{preq}}_{\theta_0}(y_{1:n}|x_{1:n}) + \sum_{i=1}^n\log_2p_M(y_i|x_i) \label{eqn:km}
\end{align}
The conditional complexity of dataset $D$ after observing model $M$ is measured with the cross-entropy of $M$ on $D$. This is the remaining information in $D$ that is not encoded in the weights of $M$.

Prior work, mainly \cite{NIPS2018_7490} and \cite{DBLP:journals/corr/abs-2003-12298} use (\ref{eqn:km}) to measure the learned information in $M$. However, there is a fundamental defect in this method: it measures both generalizable information and the noise simply remembered by the model. The last term of (\ref{eqn:km}) is heavily affected by model $M$'s degree of overfit. Because overfitting is common in deep learning, $K(M)$ is not directly linked with generalization.

If we are interested in the ``knowledge'' (or patterns in the data) learned by the model, we need to use a different formulation, which we denote by $L_{model}$:
\begin{align}
	L_{model}(M) &=: K(M) - K(M|D^*) \label{equ:definition}\\
	&= K(D^*) - K(D^*|M)\\
	&\approx L^{\text{preq}}_{\theta_0}(y^*_{1:k}|x^*_{1:k}) - L^{\text{preq}}_M(y^*_{1:k}|x^*_{1:k})
\end{align}

A separate dataset $D^*$ (for example, the test set) is used to measure the codelength. We subtract $K(M|D^*)$ from $K(M)$ to get the generalizable information in $M$. $K(M|D^*)$ measures the noise $M$ remembered from $D$. By ``noise'' we mean patterns that only exist in one particular sample. If we let $D^*$ be infinitely large (but does not include examples in $D$), then $K(M|D^*)$ is exactly the noise in $M$ (the part of $M$ that cannot be determined even when knowing all patterns in the task). But in practice $D^*$ is usually of limited size, so we propose a practical measure $L_{IT}^k$ as below ($^*$ is omitted)
\begin{align}
	L_{IT}^k(M) = L^{\text{preq}}_{\theta_0}(y_{1:k}|x_{1:k}) - L^{\text{preq}}_M(y_{1:k}|x_{1:k}) 
\end{align}

%
Intuitively, we measure the information a model gained during training by comparing the codelength of encoding $k$ examples (not overlapping with training set) with the model before ($\theta_0$) and after training ($M$). The more information the model gains about the task from training, a larger reduction of codelength would be observed. We call it \textit{information transfer} because it is correlated with the amount of generalizable information transferred from the dataset to the model during training. 

The above formulation also applies when the coding set $D^*$ and the training set $D$ are drawn from different tasks. If $D^*$ is drawn from task $T^*$ and $D$ is drawn from task $T$, then by the definition of (\ref{equ:definition}) $L_{IT}$ measures the information in $M$ about task $T^*$. $L_{IT}>0$ when $T^*$ and $T$ have common patterns.

%

\subsection{Properties of $L_{IT}$}
In the definition of $L_{IT}$, $k$ can be treated as an auxiliary variable that is independent of the size of the training set. For all possible $k$, the smallest is $k=1$:
\begin{align*}
	&L_{IT}^1(M) = -\log p_{\theta_0}(y_1|x_1) + \log p_M(y_{1}|x_{1})\\
	&\mathbb{E}[L_{IT}^1(M)] = \mathcal{L}_{val}(\theta_0) - \mathcal{L}_{val}(M)
\end{align*}
When the loss function $\mathcal{L}$ is cross-entropy, the expectation of $L_{IT}^1$ reduces to a conventional generalization performance measure: the reduction of validation loss.

On the other hand, if $k\rightarrow \infty$: (Proof of (\ref{equ:k_inf}) and (\ref{equ:bound}) is given in Appendix \ref{appendix:a}.)
\begin{align}
	L_{IT}^\infty(M) \leq -\sum_{i=1}^n\log p_{\theta_{i-1}}(y_i|x_i) + \sum_{i=1}^n\log p_{\theta_{oracle}}(y_i|x_i)
\label{equ:k_inf}
\end{align}
where $\theta_{oracle}$ is an oracle model in the model family, $\{x_{1:n},y_{1:n}\}$ is $M$'s training set. Because $L_{IT}^k$ is monotonic increasing with $k$, this tells us that $L_{IT}^k(M)$ is bounded for every $k$:
\begin{align*}
	L_{IT}^1(M) \leq L_{IT}^k(M) \leq L_{IT}^\infty(M)
\end{align*} 

We can show that the theoretical complexity of generalizable information in (\ref{equ:definition}) also satisfy:
\begin{align}
		K(M) - K(M|D^*) \leq -\sum_{i=1}^n\log p_{\theta_{i-1}}(y_i|x_i) + \sum_{i=1}^n\log p_{\theta_{oracle}}(y_i|x_i)
\label{equ:bound}
\end{align}

\begin{figure}[t]
\begin{subfigure}{.33\textwidth}
  \centering
  \includegraphics[width=\linewidth]{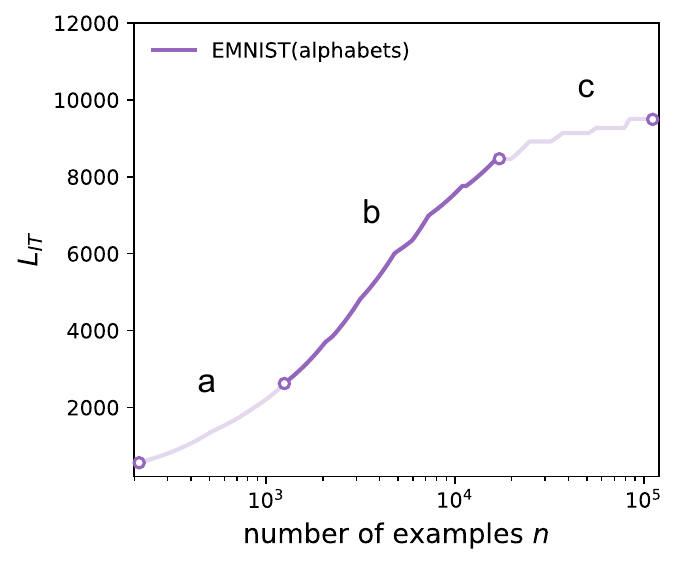}
\end{subfigure}%
\begin{subfigure}{.33\textwidth}
  \centering
  \includegraphics[width=\linewidth]{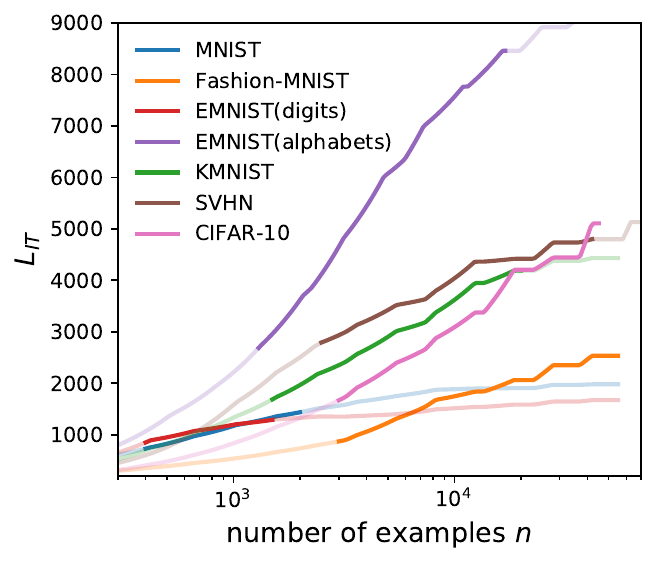}
\end{subfigure}
\begin{subfigure}{.33\textwidth}
  \centering
  \includegraphics[width=\linewidth]{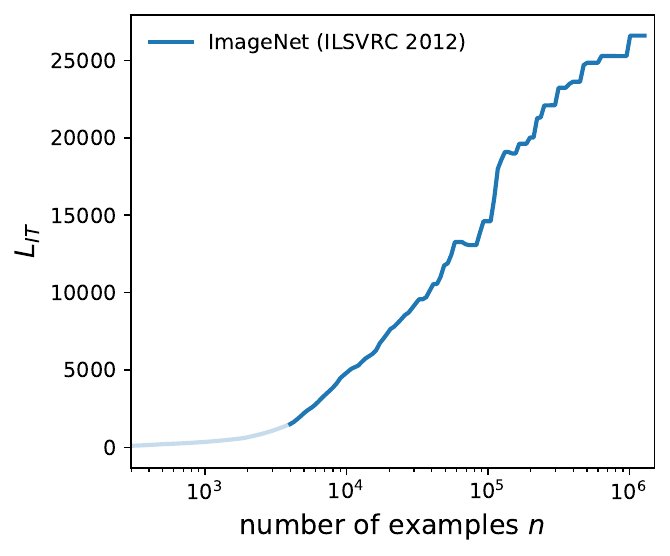}
\end{subfigure}
\caption{$L_{IT}$ as a function of $n$. The number of examples $n$ is in log scale. Left: The three phases of $L_{IT}(\theta_n)$ curve. Middle and right: Small and large datasets; the linear relationship is highlighted.}
\label{fig:wrtn}
\end{figure}

Next we investigate the behavior of $L_{IT}(\theta_n)$ for varying $n$. Intuitively, the generalizable information in model $\theta_n$ grows as the number of training examples $n$ increases, with diminishing growth when $n$ is large enough. In Figure \ref{fig:wrtn}, we plot $L_{IT}(\theta_n)$ as a function of $n$ for a number of datasets. A common pattern is observed: after a short initiation phase, $L_{IT}$ grows roughly linearly with $\log n$, and  finally saturates. The linear growing phase is where learning primarily takes place, and saturation happens when the model is close to converging, as the model stops gaining new information from more data.

In the MDL literature, \cite{grunwald2010prequential} pointed out that for a variety of model families, the codelength of the model is proportional to $\log n$ (the second right-hand term in (\ref{equ:connection})):
\begin{align}
	L^{\text{preq}}_\theta(x_{1:n}) = - \sum_{i=1}^n\log_2p_{\theta_n}(x_i) + \frac{d}{2}\log n + O(1) 
\label{equ:connection}
\end{align}

Although neural networks are much more complex than those simple model families, the log-linear relationship between model information estimated by $L_{IT}$ and the training set size $n$ is similar. 





\section{Measuring Information Transfer in Neural Networks}

In this section, we apply the information transfer $L_{IT}$ in experiments to see that it can serve as a reasonable measure of generalizable information in a model. Furthermore, $L_{IT}$ can be used to measure generalizable information in a dataset, which is lower-bounded by model information. This makes $L_{IT}$ a useful analytic tool to understand models and datasets.

As discussed in Section \ref{sec:lit}, an important property of $L_{IT}$ is that it only measures generalizable knowledge. Information gain from fitting (or remembering) particular examples does not contribute to $L_{IT}$ (see Appendix \ref{appendix:remember} for an example). Although the absolute value of $L_{IT}$ depends on the choice of hyper-parameter $k$, by fixing $k$ to a certain value we are able to compare the information content in different models and in different datasets. We use $k=5000$ for small datasets (MNIST variants and CIFAR-10) and $k=10000$ for other datasets throughout our experiments. More experiment details that are not elaborated here can be found in the appendix.

\subsection{Synthetic Dataset}
First we verify the correlation between $L_{IT}$ and $L_{model}$ using synthetic datasets. We generate text corpus using 2-gram language models, then we train LSTM language models on them. The amount of generalizable information ($L_{model}$) can be controlled by varying the number of independently sampled 2-grams in the model generating the corpus. We measure $L_{IT}$ of converged LSTM language models. In Figure \ref{fig:syn} we observe a roughly linear relationship between $L_{IT}$ and $L_{model}$ ($\rho\approx 0.98$).

\begin{minipage}{\textwidth}
  \begin{minipage}[t][][t]{0.41\textwidth}
    \centering
    \includegraphics[trim={0 0 0 0.2cm},clip,width=0.95\linewidth]{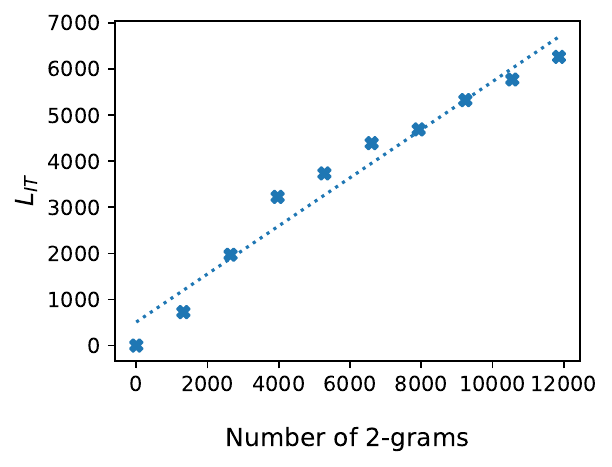}
    \captionof{figure}{Relationship between $L_{IT}$ and $L_{model}$ on synthetic corpus.}
    \label{fig:syn}
  \end{minipage}
  \hfill
  \begin{minipage}[t][][t]{0.55\textwidth}
    \centering
    \includegraphics[width=0.9\linewidth]{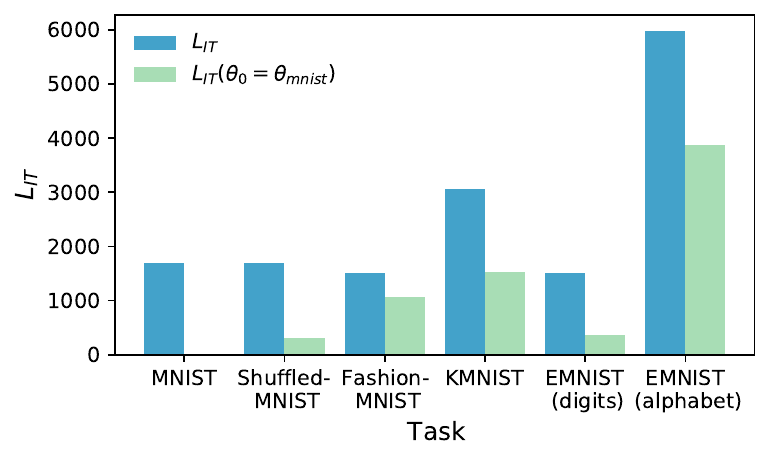}
    \captionof{figure}{$L_{IT}$ on MNIST-variants. $L_{IT}(\theta_0=\theta_{mnist})$ measures information transfer when initialized from MNIST}
    \label{fig:mnist}
    \end{minipage}
  \end{minipage}
  
\subsection{Measuring Information in MNIST-variants}
\label{sec:mnist}
Next we apply $L_{IT}$ to a ResNet-18 model on the commonly used MNIST dataset and its variants. Figure \ref{fig:mnist} illustrates $L_{IT}$ of a model trained on each dataset. One is able to tell that a model classifying 10 Japanese characters (KMNIST \cite{clanuwat2018deep}) contains significantly more information than a model classifying 10 digits (MNIST \cite{lecun1998gradient} and EMNIST \cite{cohen2017emnist}). Because Japanese characters are more complex than digits, it requires more knowledge to classify the former. A model classifying the English alphabet (upper and lower-case, 47 classes \cite{cohen2017emnist}) contains even more information, as the task is even more difficult.

The model information measured with $L_{IT}$ is relative to the initial model $\theta_0$. Therefore it gives the information gain, or ``new knowledge'' learned from the dataset compared to what is already in the initial model $\theta_0$. We can use a model pre-trained on MNIST ($\theta_{mnist})$ as $\theta_0$ to measure the amount of new information a model can learn from each dataset. From $L_{IT}(\theta_0=\theta_{mnist})$ in Figure \ref{fig:mnist}, KMNIST, Fashion-MNIST \cite{xiao2017/online} and EMNIST-alphabet all contain much new information that is not in MNIST. Shuffled-MNIST is a version of MNIST with its labels permuted, therefore it only introduces a little new information. The same is with EMIST-digits, which only differ from MNIST in image pre-processing.

Comparing $L_{IT}(\theta_0=\theta_{mnist})$ with $L_{IT}$, knowledge about MNIST significantly reduces new information gained from Japanese characters and English alphabet, signifying these datasets share some common information with MNIST (for example, stroke patterns). On the other hand, $L_{IT}$ on Fashion-MNIST does not reduce as much, showing there is less information in common between handwritten digits and clothes images.

\subsection{Dissecting Information in CIFAR-10}
\label{sec:dissect}
In the following example, we use $L_{IT}$ to dissect the knowledge about object classification in CIFAR-10, which also shows the consistency of $L_{IT}$ in measuring information.

\begin{figure}
  \centering
  \includegraphics[width=0.7\linewidth]{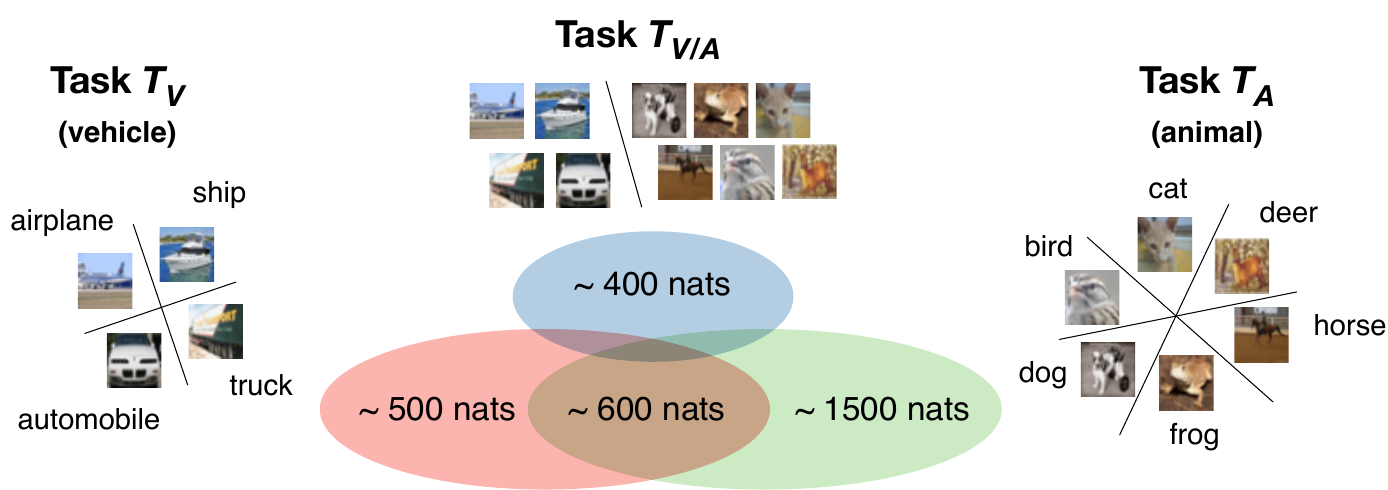}
  \caption{Dissecting information in CIFAR-10}
  \label{fig:dissect_cifar}
\end{figure}

 The 10 object classes in CIFAR-10 belong to two categories: vehicles (4 classes) and animals (6 classes). We can therefore split CIFAR-10 into three tasks: $T_V$ and $T_A$ for classifying within vehicles and animals respectively, and $T_{V/A}$ for classifying between the two categories. By measuring $L_{IT}$ on learning each subtask, and on transfer learning from one subtask to another, we are able to produce a Venn diagram representing the information content and the relationship of three subtasks (Figure \ref{fig:dissect_cifar}).

\begin{figure}[h]
\begin{subfigure}{.47\textwidth}
  \centering
  \includegraphics[width=\linewidth]{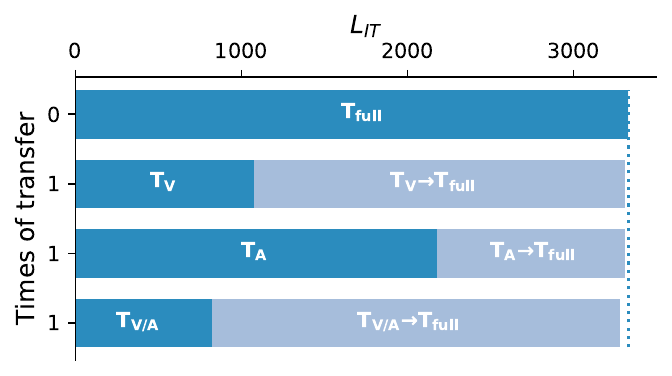}
\end{subfigure}%
\hskip .05\textwidth
\begin{subfigure}{.47\textwidth}
  \centering
  \includegraphics[width=\linewidth]{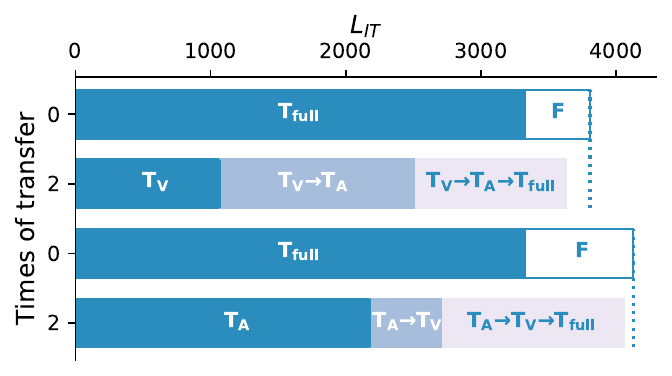}
\end{subfigure}
\caption{$L_{IT}$ measured on models sequentially learning subtasks of CIFAR-10. $F$ in the right plot measures the information forgotten about the first task in transferring to a second task (See also details in Appendix \ref{appendix:cifar})}
\label{fig:cifar}
\end{figure}

The information about classifying objects can be learned from the whole CIFAR-10 dataset $T_{full}$ as well as from a combination of subtasks. Figure \ref{fig:cifar} plotted $L_{IT}$ measured on sequentially training on subtasks; for example, training on $T_V$ first, then on $T_{A}$ ($T_V$\textrightarrow $T_{A}$). $L_{IT}$ always adds up roughly to a fixed value (the total information of CIFAR-10), regardless of how one learns from subtasks. This is further evidence that $L_{IT}$ is consistently correlated with the true amount of information.

$L_{IT}$ gives us a lot of information about the task CIFAR-10. For example, classifying animals requires much more information than classifying vehicles. In the knowledge used to classify animals, about $1/4$ can be shared with classifying vehicles, and the remaining $3/4$ is specific to animals. $L_{IT}$ also reveals the dynamics of neural network transfer learning. For example in experiments involving a second transfer, one is able to tell from $L_{IT}$ the amount of information forgotten about the previous task, after training the model on a new task.

\section{Measuring Information Transfer in Transfer Learning}
The success of transfer learning depends on sharing knowledge from previous tasks \cite{pan2009survey}. Recently we have seen great success with pre-trained representation  models in vision and NLP. Models pre-trained on large-scale datasets acquire good general knowledge of image and text, which helps them achieve state-of-the-art performance when adapted to a variety of tasks.

A natural question then is how to measure the quality of pre-trained representation models. Better representations not only achieve better performance on target tasks, but also require fewer training examples to reach a certain level of performance. As pointed out by \cite{yogatama2019learning}, the ability to generalize rapidly to a new task by using previously acquired knowledge cannot be measured by performance metrics alone. They proposed to use prequential codes as a measure of linguistic intelligence. 

From an informational perspective, the quality of representation models can be measured by the amount of generalizable information in the model, which can be reused in downstream tasks. 
To facilitate comparison of different models using information transfer, we introduce the \textit{Information Advantage} of model $M$ over $M_{r}$:
\begin{align}
	L_{IA}(M,M_r) &= L_{IT}(M) -  L_{IT}(M_r)\\
	&= L^{\text{preq}}_{M_{r}}(y_{1:k}|x_{1:k}) - L^{\text{preq}}_{M}(y_{1:k}|x_{1:k}) 
\end{align}
If the reference model $M_r$ is taken to be a random model $\theta_0$, then $L_{IA}$ becomes $L_{IT}$. $L_{IA}$ measures how much information $M$ has more than $M_{r}$ about the target dataset $\{x,y\}$. 




In Table \ref{tab:trans}, we measure the model information of pre-trained image classification models and pre-trained language models. The target tasks are CIFAR-10 and MultiNLI. Examining model performance, clearly there is not a single measure that represents the quality of the model: Different models have an advantage at different numbers of training examples. Task-specific knowledge (from pre-training on a similar task) contributes to few-shot effectiveness, while general knowledge helps more in many-shot. Both kinds of knowledge can be measured and are reflected in $L_{IA}$.

More expressive model and pre-training on more data both contribute to increased  information in the model by $L_{IA}$. Pre-training on similar tasks (e.g., from SNLI to MultiNLI) is also a method to significantly increase model information about the target task. From Table \ref{tab:trans} we can observe that if a model have more information about the target task, it will generally have better performance across different transfer learning settings.

\begin{table}[t]
\setlength{\abovecaptionskip}{7pt}
  \fontsize{8}{9}\selectfont
  \caption{Information transfer in transfer learning. AlexNet and BERT-base are used as the reference models in $L_{IA}$. De-CIFAR stands for a color-distorted version of CIFAR-10. \textcolor{red}{Red} color marks the top-2 best performances. \textbf{Bold} marks the top-2 in model information}
  \label{tab:trans}
  \centering
  \begin{tabular}{lcrrrr}
  \toprule
  CIFAR-10 \cite{krizhevsky2009learning} \\
   & & Information & \multicolumn{3}{c}{Performance (Accuracy \%)} \\
   \cmidrule(r){3-3} \cmidrule(r){4-6}
  Model & Pretraining & $L_{IA}$\footnotemark & Zero-shot & Few-shot{\tiny$(10^2)$} & Many-shot{\tiny$(10^4)$}\\
  \midrule
  AlexNet \cite{krizhevsky2014one}   & ImageNet   \cite{imagenet_cvpr09}    & 0    &    10.0    & 47.1  & 86.7 \\
  VGG11 \cite{simonyan2014very}    & ImageNet                 & 0.55  &    10.0    & 48.7  & 87.9 \\
  ResNet-18 \cite {he2016deep} & CIFAR-100 \cite{krizhevsky2009learning}       & -0.49 &    10.0    & 46.6  & 82.6 \\
  ResNet-18 & De-CIFAR              & 1.33 &    \textcolor{red}{70.8}    & \textcolor{red}{71.9}  & 84.6 \\
  ResNet-18 & ImageNet   			  & 2.44 &    10.0    & 58.9  & 91.8 \\
  ResNet-18 & ImageNet + De-CIFAR   & \textbf{4.05} &    \textcolor{red}{86.6}    & \textcolor{red}{87.9}  & \textcolor{red}{93.3} \\   
  ResNet-34 & ImageNet                & 2.64 &    10.0    & 61.7  & 92.9 \\
  ResNeXt-50 \cite{xie2017aggregated} & ImageNet               & \textbf{3.07} &    10.0    & 62.5  & \textcolor{red}{93.1} \\
  \midrule
  MultiNLI \cite{N18-1101}\\
   & & Information & \multicolumn{3}{c}{Performance (Accuracy \%)} \\
   \cmidrule(r){3-3} \cmidrule(r){4-6}
  Model & Pretraining & $L_{IA}$ & Zero-shot & Few-shot{\tiny$(10^3)$} & Many-shot{\tiny \medmuskip=0 mu $(5\cdot10^4)$} \\
  \midrule
  BERT-base \cite{devlin-etal-2019-bert}      & Unsup.      & 0     & 33.3  & 50.9  & 76.4 \\
  BERT-base      & Unsup. + SNLI \cite{snli:emnlp2015} & 1.44  & \textcolor{red}{65.9}  & 69.6  & 78.2 \\
  BERT-large     & Unsup.        & 0.79   & 33.3  & 61.5  & 79.8 \\
  XLNet-large \cite{yang2019xlnet}    & Unsup.        & \textbf{2.30}  & 33.3  & \textcolor{red}{76.9}  & 84.7 \\
  RoBERTa-large \cite{liu2019roberta}  & Unsup.        & 1.90  & 33.3  & 69.2  & \textcolor{red}{86.0} \\  
  RoBERTa-large  & Unsup. + SNLI &   \textbf{3.51}  & \textcolor{red}{78.7}  & \textcolor{red}{81.6}  & \textcolor{red}{85.8} \\  
  \bottomrule
  \end{tabular}
 \end{table}
\footnotetext{from hereafter, $L_{IT}$ and $L_{IA}$ are measured in $k$-$nats$}

\section{Measuring Information Transfer in Continual Learning}
\label{sec:cont}
The ability to continually learn is a fundamental element of genuine intelligence. Neural network models, although powerful, tend to suffer from catastrophic forgetting \cite{mccloskey1989catastrophic,french1999catastrophic} in continual learning settings. Some methods have been proposed (for example, \cite{kirkpatrick2017overcoming,lee2017overcoming}) to prevent the model from forgetting about old tasks. However, recent studies pointed out that these methods often fail to significantly prevent forgetting or being unfeasible in practice \cite{kemker2018measuring,pfulb2019comprehensive}.

We take a different route to investigate catastrophic forgetting, by estimating model information with $L_{IA}$: ``Forgetting,'' after all, means losing information. We found that information tells quite a different story than performance metrics in continual learning. 

Experiments are performed in two scenarios: \textit{image classification}, where we split the 200-class Tiny-ImageNet 
 into four 50-class classification tasks, and \textit{language modeling}, where we extracted 4 topics from the 1-billion-word-benchmark corpus \cite{ChelbaMSGBKR14} as four tasks. Algorithms employed are plain transfer, L2 regularization, Elastic Weight Consolidation \cite{kirkpatrick2017overcoming}, and Incremental Moment Matching (weight-transfer and L2-transfer) \cite{lee2017overcoming}. Single-task and multi-task training are used as baselines. Models used are ResNet-56 for image classification and 2-layer LSTM for language modeling.

In Table \ref{tab:cont}, we compare performance metrics as well as $L_{IA}$.  $L_{IA}$ measures the amount of information the final model has about each task. Under \textit{All past} is given the performance and information on all four past tasks combined. In \textit{Future} we transfer the final model to a larger task to examine the representation learned throughout continual learning. We have the following observations:

	

\begin{table}[h]
\setlength{\abovecaptionskip}{7pt}
  \fontsize{8}{9}\selectfont
  \caption{Information transfer in continual learning. \textit{acc.} and \textit{ppl.} stand for \textit{accuracy} and \textit{perplexity}. \textcolor{red}{Red} color marks the top-2 best performance. \textbf{Bold} marks the top-2 model information.}
  \label{tab:cont}
  \centering
  \begin{tabular}{lrrrrrrrrrrrr}
    \toprule
    Tiny-ImageNet \\
     & \multicolumn{2}{c}{Task 0} & \multicolumn{2}{c}{Task 1} & \multicolumn{2}{c}{Task 2} & \multicolumn{2}{c}{Task 3} & \multicolumn{2}{c}{All past} & \multicolumn{2}{c}{Future}\\
     \cmidrule(r){2-3} \cmidrule(r){4-5} \cmidrule(r){6-7} \cmidrule(r){8-9} \cmidrule(r){10-11} \cmidrule(r){12-13}
    Method     & acc. & $L_{IA}$    & acc. & $L_{IA}$    & acc. & $L_{IA}$    & acc. & $L_{IA}$    & acc. & $L_{IA}$   & acc. & $L_{IA}$ \\
    \midrule
       plain      &   7.1   &    8.4   & 17.8  &      9.1    & 25.5  &   10.5    & 61.1  &  13.9   & 27.9  & \textbf{41.8} &      \textcolor{red}{\textbf{33.9}}    & \textbf{7.6} \\
    L2         &   44.2  &    11.3  & 38.2  &      9.5    & 37.8  &   9.1     & 45.6  &  9.0    & 41.5  & 38.8 &      30.7    & 6.0 \\
    EWC        &   47.9  &    11.5  & 43.5  &      9.2    & 41.6  &   9.1     & 44.8  &  8.9    & 44.5  & 38.6 &      29.7    & 5.4 \\
 IMM-mean (wt) &   27.9  &    10.7  & 45.0  &      11.9   & 41.5  &   11.9    & 41.2  &  10.3   & 38.9  & \textbf{44.8} &        \textcolor{red}{\textbf{32.8}}  & \textbf{7.7} \\
 IMM-mode (wt) &   12.1  &    9.3   & 25.2  &      11.3   & 27.3  &   11.0    & 21.7  &  9.9    & 21.6  & 41.4 &        32.4  & 6.8 \\
 IMM-mean (l2) &   57.7  &    12.0  & 50.6  &      9.7    & 49.1  &   9.7     & 48.7  &  9.5    & \textcolor{red}{51.5}  & 40.9 &        28.9  & 5.6 \\
 IMM-mode (l2) &   57.4  &    12.1  & 52.2  &      9.6    & 49.6  &   9.3     & 47.9  &  9.1    & \textcolor{red}{51.8}  & 40.1 &        28.6  & 5.5 \\
 \midrule
   Single-task &   56.6  &       13.2 & 60.7  &        12.8 & 56.3  &        13.3 & 53.9  &      12.7       &    -   & -     &        -      & - \\
    Multi-task &   62.8  &    14.7  & 63.6  &      14.4   & 61.1  &   14.3    & 61.1  &  13.9   & 62.2  & 57.3 &        38.6  & 9.8 \\
	\midrule 
	\multicolumn{2}{l}{1b-word-benchmark}\\
     & \multicolumn{2}{c}{Task 0} & \multicolumn{2}{c}{Task 1} & \multicolumn{2}{c}{Task 2} & \multicolumn{2}{c}{Task 3} & \multicolumn{2}{c}{All past} & \multicolumn{2}{c}{Future}\\
     \cmidrule(r){2-3} \cmidrule(r){4-5} \cmidrule(r){6-7} \cmidrule(r){8-9} \cmidrule(r){10-11} \cmidrule(r){12-13}
    Method     & ppl. & $L_{IA}$    & ppl. & $L_{IA}$    & ppl. & $L_{IA}$    & ppl. & $L_{IA}$    & ppl. & $L_{IA}$   & ppl. & $L_{IA}$ \\
    \midrule
    plain      &   283  &   20.6& 395   & 19.8& 370       & 24.8& 193     & 29.5&  300   & \textbf{94.6}&   \textcolor{red}{\textbf{256}}   & \textbf{23.8} \\
    L2         &   213  &   23.7& 298   & 20.5& 381       & 20.1& 251     & 24.0& \textcolor{red}{279}   & 88.2&        281   & 22.7 \\
    EWC        &   197  &   23.2& 233   & 22.4& 300       & 22.7& 262     & 23.3& \textcolor{red}{244}   & 91.6&        304   & 21.0 \\
 IMM-mean (wt) &   437  &   22.8& 399   & 24.4& 687       & 25.7& 1752    & 20.0& 666    & \textbf{92.9}&   \textcolor{red}{\textbf{258}}   & \textbf{24.1} \\
 IMM-mode (wt) &   288  &   23.2& 341   & 24.2& 772       & 23.0& 1854    & 17.1& 605    & 87.5&        289 & 20.3 \\
   \midrule
   Single-task &  111   &        28.7 & 109   &        30.0 & 152         & 31.1 & 232         & 26.7 &  -     & -   &        -   & -     \\
    Multi-task &   90   &   31.2& 90    & 31.7& 126       & 32.8& 203     & 29.4&  119   & 125&        202 & 29.7 \\
    \bottomrule
  \end{tabular}
\end{table}


\begin{figure}[h]
  \centering
  \begin{subfigure}{.33\textwidth}
  \centering
  \includegraphics[width=\linewidth]{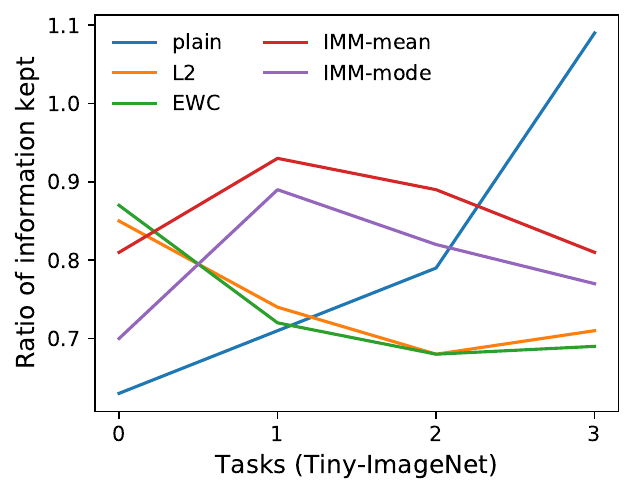}
\end{subfigure}\hspace{13mm}
\begin{subfigure}{.33\textwidth}
  \centering
  \includegraphics[width=\linewidth]{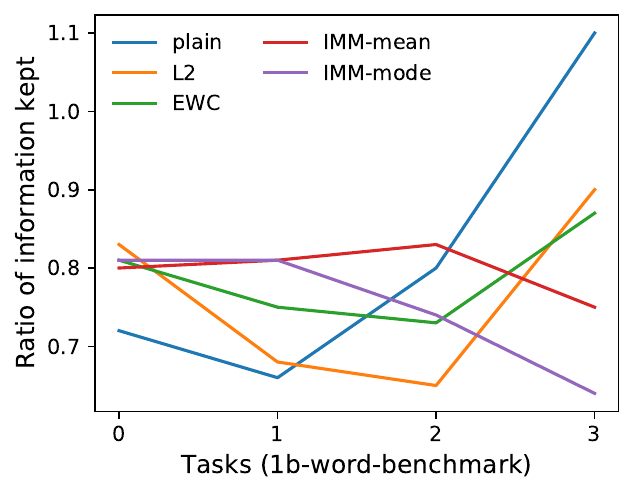}
\end{subfigure}
  \caption{Ratio of information kept about each task, in the final model.}
  \label{fig:ratio}
\end{figure}

\textbf{$L_{IA}$ measures forgetting while performance alone does not}. We found that performance on past tasks is largely irrelevant to the performance on future tasks. Model information (as measured by $L_{IA}$), however, is correlated with both performance and model information on future tasks. This indicates that $L_{IA}$ is a reasonable measure of generalizable knowledge in the model: Models with larger $L_{IA}$ preserve more information about past tasks, thus transfer more effectively to new tasks.

\begin{figure}[h]
  \centering
  \includegraphics[width=0.6\linewidth]{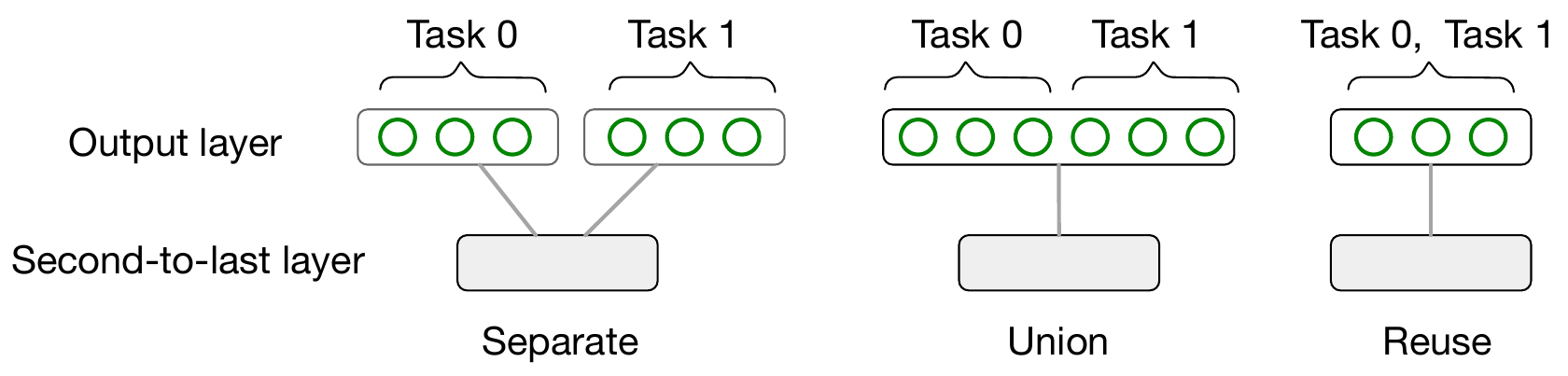}
  \caption{Three common situations of the final output layer in continual learning}
  \label{fig:paradigms}
\end{figure}

$L_{IA}$ can also help us understand why measuring forgetting with performance alone is flawed. Figure \ref{fig:paradigms} shows three common strategies to deal with the final output layer in continual learning. Using EWC in the three situations yields very different results in performance (Table \ref{tab:paradigms}). Only ``separate'' is able to perform well, because it keeps the final layer of past tasks intact.
However, from $L_{IA}$, one is able to tell that the ``separate'' and the ``union'' strategies are very similar in the ability to preserve information. ``Reuse'' is less effective because the shared final layer impedes learning.


\begin{table}[b]
	\setlength{\abovecaptionskip}{7pt}
  \fontsize{8}{9}\selectfont
  \caption{Continual learning using EWC, with the three strategies}
  \label{tab:paradigms}
  \centering
  \begin{tabular}{lrrrrrrrrrr}
    \toprule
    & \multicolumn{2}{c}{Task 0} & \multicolumn{2}{c}{Task 1} & \multicolumn{2}{c}{Task 2} & \multicolumn{2}{c}{Task 3} & \multicolumn{2}{c}{All past}\\
    \cmidrule(r){2-3} \cmidrule(r){4-5} \cmidrule(r){6-7} \cmidrule(r){8-9} \cmidrule(r){10-11}
    Method     & acc. & $L_{IA}$    & acc. & $L_{IA}$    & acc. & $L_{IA}$    & acc. & $L_{IA}$    & acc. & $L_{IA}$ \\
    \midrule
    Separate & 47.9  & 11.5 & 43.5 & 9.2  & 41.6 & 9.1  & 44.8  & 8.8  & 44.5 & \textbf{38.6} \\
    Union    & 0.0   & 10.4 & 0.0  & 8.9  & 0.0  & 8.8  & 45.6  & 9.4  & 11.4 & \textbf{37.6} \\
    Reuse    & 4.3   & 9.9  & 3.4  & 6.8  & 3.1  & 5.8  & 41.3  & 6.4  & 13.0 & 28.8 \\    
	\bottomrule
  \end{tabular}
\end{table}


\textbf{Catastrophic forgetting in neural networks is not really catastrophic}. From the above discussion, we argue that forgetting needs to be understood at a deeper level. Forgetting involves two distinct factors: preserving knowledge and preserving the exact decision boundary. We use $L_{IA}$ to measure the former factor. While plain transfer fails by performance metrics, it does a decent job of preserving knowledge, keeping over $70\%$ of total information in past tasks. This is also confirmed by its performance on future tasks (which is not affected by the exact decision boundary of past tasks).

\textbf{Information transfer provides a new perspective on the effectiveness of continual learning.} Continual learning strives to make a model learn like a human and to become more intelligent over time. Therefore, there are two main goals of continual learning: 1) perform well on past tasks (not forgetting), and 2) adapt quickly and perform well on new tasks (learn better representations). Previous works mainly focus on the first goal. Either goal requires the ability to preserve information, while the second goal is more directly determined by it.

In comparing the effectiveness of preserving information in Table \ref{tab:cont}, no continual learning method is significantly more effective than plain transfer learning. Figure \ref{fig:ratio} plots the ratio of preserved information on each task, and we can see that each method achieves a different balance of information: Regularization-based methods (L2, EWC) preserve more information from the first and the last tasks, while model averaging (IMM) more favors information from tasks in the middle. Plain transfer clearly remembers most about the last task. It is an interesting phenomenon that no method is able to cover the significant gap between continual learning and multi-task learning, given that neural networks are often over-parameterized \cite{zhang2016understanding} and have more than enough capacity. 

\section{Conclusion}

In this paper, we have presented a practical measure of generalizable information in a neural network model, and showed in multiple scenarios that model information and dataset information are important and useful metrics for analyzing deep learning problems. Measuring exact information in a neural network is a non-trivial task, but an approximate measure such as the proposed \textit{information transfer} can have good correlation with true model information and is easy to calculate. We hope information transfer serves as a tool to motivate further investigation of neural models and their behaviors in learning from an information perspective. It is also a relevant future direction to develop learning algorithms to maximize generalizable model knowledge.

%
%
%
%
%

\bibliographystyle{unsrt}
\bibliography{if.bib}

\clearpage

\appendix
\section*{Appendix}
\section{The Information Transfer Measure}
\label{appendix:a}

\subsection{Proof of Equation (\ref{equ:k_inf})}

From the definition of $L_{IT}$, as $k\rightarrow\infty$
\begin{align*}
L_{IT}^k(\theta_n) &= L^{\text{preq}}_{\theta_0}(y^*_{1:k}|x^*_{1:k}) - L^{\text{preq}}_{\theta_n}(y^*_{1:k}|x^*_{1:k}) \\
&\leq L^{\text{preq}}_{\theta_0}(y^*_{1:k},y_{1:n}|x^*_{1:k},x_{1:n}) - L^{\text{preq}}_{\theta_{oracle}}(y_{1:n}|x_{1:n}) - L^{\text{preq}}_{\theta_n}(y^*_{1:k}|x^*_{1:k}) \\
&= L^{\text{preq}}_{\theta_0}(y_{1:n},y^*_{1:k}|x_{1:n},x^*_{1:k}) - L^{\text{preq}}_{\theta_{oracle}}(y_{1:n}|x_{1:n}) - L^{\text{preq}}_{\theta_n}(y^*_{1:k}|x^*_{1:k}) \\
&\leq L^{\text{preq}}_{\theta_0}(y_{1:n}|x_{1:n}) + L^{\text{preq}}_{\theta_n}(y^*_{1:k}|x^*_{1:k}) - L^{\text{preq}}_{\theta_{oracle}}(y_{1:n}|x_{1:n}) - L^{\text{preq}}_{\theta_n}(y^*_{1:k}|x^*_{1:k}) \\
&= L^{\text{preq}}_{\theta_0}(y_{1:n}|x_{1:n}) - L^{\text{preq}}_{\theta_{oracle}}(y_{1:n}|x_{1:n})
\end{align*}
assume model converges to $\theta_{oracle}$ given a sufficiently large dataset. The above derivation makes the assumption that when the coding set is sufficiently large, $L^{\text{preq}}$ is independent of the order of coding. Therefore we have
\begin{align*}
	L_{IT}^\infty(M) \leq -\sum_{i=1}^n\log p_{\theta_{i-1}}(y_i|x_i) + \sum_{i=1}^n\log p_{\theta_{oracle}}(y_i|x_i)
\end{align*}

\subsection{Proof of Equation (\ref{equ:bound})}

\begin{align*}
	K(M) - K(M|D^*) &= K(D) - K(D|M) - K(M|D^*)\\
	&\leq K(D) - K(D|M) - K(M|\theta_{oracle})\\
	&\leq K(D) - K(D|\theta_{oracle})\\
	&= K(D) + \sum_{i=1}^n\log p_{\theta_{oracle}}(y_i|x_i) \\
	&\leq -\sum_{i=1}^n\log p_{\theta_{i-1}}(y_i|x_i) + \sum_{i=1}^n\log p_{\theta_{oracle}}(y_i|x_i)
\end{align*}

\subsection{Connection Between $L_{IT}$ and Performance Metrics}
For classification tasks with cross-entropy loss function, the prequential coding curve (codelength per example as a function of example index) is very close to the ``validation loss - dataset size'' curve (Figure \ref{fig:app_Landval}):

\begin{figure}[h]
  \centering
  \includegraphics[width=0.5\linewidth]{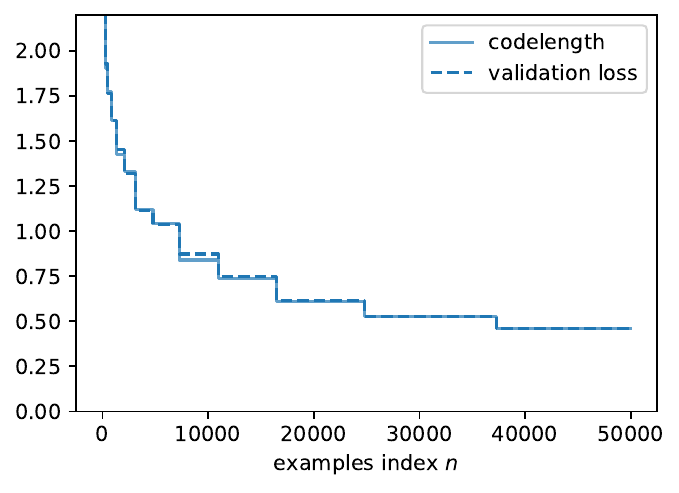}
  \caption{The prequential coding curve and the ``validation loss - dataset size'' curve on CIFAR-10}
  \label{fig:app_Landval}
\end{figure}

Therefore, $L^{\text{preq}}$ can be interpreted as an integration of the ``validation loss - dataset size'' curve. Especially, $L_{IA}$ can be interpreted as comparing ``how fast'' the validation loss deceases as we increase dataset size $n$. Faster decease of the validation loss symbolizes the model has more prior knowledge about the task.

\begin{figure}[h]
  \centering
  \includegraphics[width=0.5\linewidth]{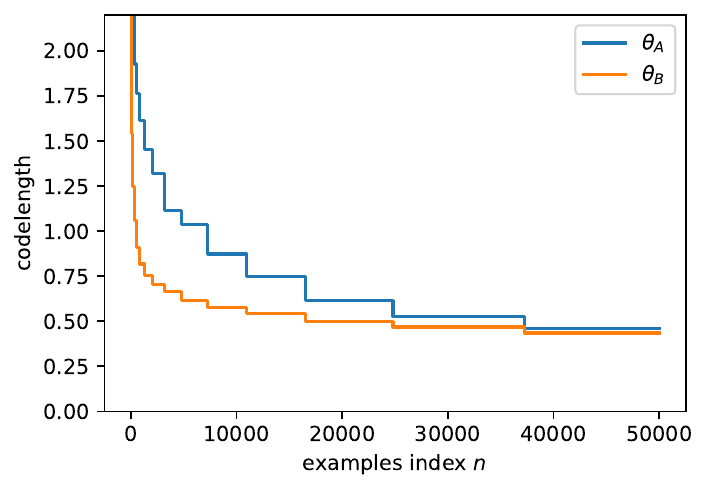}
  \caption{The prequential coding curve of two different models on CIFAR-10, $L_{IA}(\theta_B,\theta_A)$=3070}
  \label{fig:app_compare}
\end{figure}

Compare the coding curve when using two models $\theta_A$ and $\theta_B$ as initialization on CIFAR-10 in Figure \ref{fig:app_compare}, we see that both model converge to roughly the same performance, but $\theta_B$ requires much less examples to reach a given level of performance. This indicates that $\theta_B$ has more information about the task than $\theta_A$ (in fact, in this example $\theta_B$ has been pretrained on a subtask of CIFAR-10). This can also be given by the information advantage $L_{IA}$ of $\theta_B$ over $\theta_A$.

\subsection{Remembering v.s. Generalization}
\label{appendix:remember}

For a randomly-labeled dataset, although the model could achieve high accuracy on the training set, $L_{IT}(\theta_n)$ remains close to zero for any $n$ (Figure \ref{fig:app_random}):

\begin{figure}[h]
  \centering
  \includegraphics[width=0.5\linewidth]{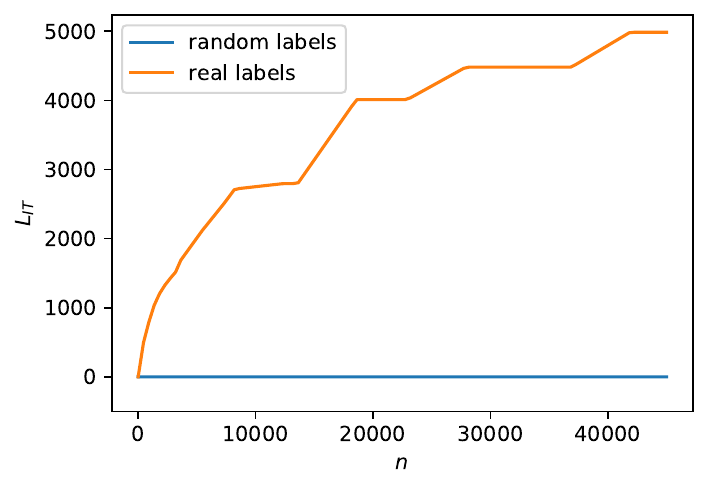}
  \caption{$L_{IT}(\theta_n)$ on CIFAR-10 with real and random labels}
  \label{fig:app_random}
\end{figure}

This is because the length of generalizable information in model $\theta_n$, $L_{model}(\theta_n)$ is zero. This shows that information transfer measure $L_{IT}(\theta_n)$ measures the generalizable information in $\theta_n$.

\section{Measuring Information Transfer in Neural Networks}

\subsection{Calculating prequential codes}
We follow the method used in \cite{NIPS2018_7490} to calculate prequential codelengths. Using the definition of $L^{preq}$ in (\ref{equ:preq}) requires training the model $n$ times, which is overly time consuming. An approximation is used in \cite{NIPS2018_7490} that first partitions the dataset, and only re-train the model on each partition. The prequential codelength becomes:
\begin{align*}
	L^{\mathrm{preq}}\left(y_{1: n} | x_{1: n}\right)=t_{1} \log K+\sum_{s=0}^{S-1}-\log p_{\hat{\theta}_{t_{s}}}\left(y_{t_{s}+1: t_{s+1}} | x_{t_{s}+1: t_{s+1}}\right)
\end{align*}
where $1=t_0 < t_1 <... <t_S = n$ is a partition of $n$ examples into $S$ sets. Because the log likelihood changes slower as the encoded example increases, we make the number of examples in each set to be 1.5 times that of the previous set. This is to make the calculation of prequential codes more time-efficient while minimizing approximation errors.
 
\subsection{Experiment Settings}
The models we used in experiments throughout this paper are ResNet-56 for image classification tasks and 2-layer LSTM of dimension 200 for language modeling tasks. Training models on the full or a subset of datasets uses early-stopping on the validation set.

\subsection{Dissecting Information in CIFAR-10}
\label{appendix:cifar}

We use $L_{IT}$ to measure the information gain in learning (or transfer learning) on a subtask of CIFAR-10. Here we use $L(T)$ to denote the amount of information transferred from task $T$ to a randomly-initialized model. $L(T\rightarrow T')$ denotes the amount of new information gained in transfer learning from task $T$ to task $T'$. It is measured by $L_{IT}(\theta)$, where $\theta$ is a model first trained on $T$ (as $\theta_0$) then transferred and trained on $T'$. $L(T\rightarrow T'\rightarrow T'')$ denotes the amount of new information gained from transferring the previous model again to a third task $T''$. $L(T\rightarrow T'\rightarrow T)$ is the information gain when transferring the model back to the first task, i.e., it measures the information forgotten about task $T$ after training on task $T'$. For example, ``F'' in Figure \ref{fig:cifar} means $L(T_V\rightarrow T_A\rightarrow T_V)$ and $L(T_A\rightarrow T_V\rightarrow T_A)$ respectively. The results are listed in Table \ref{appendix:trans-cifar}.

\begin{table}[h]
\setlength{\abovecaptionskip}{7pt}
  \small
  \caption{$L_{IT}$ on subtasks of CIFAR-10, with different initial models.}
  \label{appendix:trans-cifar}
  \centering
  \begin{tabular}{lrlrlr}
  \toprule
  \multicolumn{2}{c}{Learn from scratch} & \multicolumn{2}{c}{Transfer learning} & \multicolumn{2}{c}{Transfer twice / Re-learning} \\
  \cmidrule(r){1-2} \cmidrule(r){3-4} \cmidrule(r){5-6}
  Task & $L_{IT}$ & Task & $L_{IT}$ &Task & $L_{IT}$\\
  \midrule
  $T_V$        & 1.07 & $T_V$\textrightarrow $T_A$      & 1.45 & $T_V$\textrightarrow $T_A$\textrightarrow $T_V$       & 0.48 \\
  $T_A$        & 2.18 & $T_A$\textrightarrow $T_V$      & 0.54  & $T_A$\textrightarrow $T_V$\textrightarrow $T_A$       & 0.80 \\
  $T_{V/A}$    & 0.82 & $T_V$\textrightarrow $T_{full}$ & 2.23 & $T_V$\textrightarrow $T_A$\textrightarrow $T_{V/A}$ & 0.40 \\
  $T_{full}$   & 3.53 & $T_A$\textrightarrow $T_{full}$ & 1.13 & $T_A$\textrightarrow $T_V$\textrightarrow $T_{V/A}$ & 0.35 \\
    &  &          $T_{V/A}$\textrightarrow $T_{full}$ & 2.46 & $T_V$\textrightarrow $T_A$\textrightarrow $T_{full}$ & 1.11 \\
    &  &  &  &                                                    $T_A$\textrightarrow $T_V$\textrightarrow $T_{full}$ & 1.34 \\
  \bottomrule
  \end{tabular}
\end{table}

The following relationships of $L_{IT}$ are examined on the subtasks of CIFAR-10:
\begin{align*}
&L(T_V)+L(T_V\rightarrow T_{full})\ \equalhat\ L(T_{full}) \\
&L(T_A)+L(T_A\rightarrow T_{full})\ \equalhat\ L(T_{full}) \\
&L(T_{V/A})+L(T_{V/A}\rightarrow T_{full})\ \equalhat\ L(T_{full}) \\
&L(T_V\rightarrow T_A) + L(T_V\rightarrow T_A\rightarrow T_{full}) - L(T_V\rightarrow T_A\rightarrow T_V)\ \equalhat\ L(T_V\rightarrow T_{full}) \\
&L(T_A\rightarrow T_V) + L(T_A\rightarrow T_V\rightarrow T_{full}) - L(T_A\rightarrow T_V\rightarrow T_A)\ \equalhat\ L(T_A\rightarrow T_{full}) \\
&L(T_V\rightarrow T_A\rightarrow T_{full}) -  L(T_V\rightarrow T_A\rightarrow T_V) \ \equalhat\ L(T_V\rightarrow T_A\rightarrow T_{V/A})\\
&L(T_A\rightarrow T_V\rightarrow T_{full}) - L(T_A\rightarrow T_V\rightarrow T_A) \ \equalhat\  L(T_A\rightarrow T_V\rightarrow T_{V/A})
\end{align*}

By $\equalhat$ we mean the left-hand side approximates the right-hand side. The first three equations represent learning the task $T_{full}$ can be separated into two stages: learn information from a subtask first, then learn the rest information. The next two equations similarly represent staged learning in three stages. The final two equations represent that the information gain from $\{T_V,T_A\}$ to $T_{full}$ is exactly $T_{V/A}$.


\section{Transfer Learning}

When training data for the target task is insufficient, it is also common to use a fixed representation, and only train the final classifier layer. Some additional performance results are listed in Table \ref{appendix:trans}. The performance when only training the final layer has the similar trend as zero-shot performance. If the representation model has been trained on a similar task, performance is significantly better.

\begin{table}[h]
\setlength{\abovecaptionskip}{7pt}
  \fontsize{8}{9}\selectfont
  \caption{More performance results in transfer learning experiments. \textit{Fixed-rep} stands for fine-tuning the final classifier layer only.}
  \label{appendix:trans}
  \centering
  \begin{tabular}{lcrrrr}
  \toprule
  CIFAR-10  \\
   &  & \multicolumn{4}{c}{Performance (Accuracy \%)} \\
   \cmidrule(r){3-6}
  Model & Pretraining & Fixed-rep & Zero-shot & Few-shot{\tiny$(10^2)$} & Many-shot{\tiny$(10^4)$}\\
  \midrule
  AlexNet   & ImageNet                 & 82.7  &    10.0    & 47.1  & 86.7 \\
  VGG11     & ImageNet                 & 83.5  &    10.0    & 48.7  & 87.9 \\
  ResNet-18 & CIFAR-100                & 78.5  &    10.0    & 46.6  & 82.6 \\
  ResNet-18 & de-CIFAR-10              & 80.6  &    70.8    & 71.9  & 84.6 \\
  ResNet-18 & ImageNet   			   & 80.9  &    10.0    & 58.9  & 91.8 \\
  ResNet-18 & ImageNet + de-CIFAR-10   & 92.6  &    86.6    & 87.9  & 93.3 \\   
  ResNet-34 & ImageNet                 & 81.4  &    10.0    & 61.7  & 92.9 \\
  ResNeXt-50& ImageNet                 & 83.3  &    10.0    & 62.5  & 93.1 \\
  \midrule
  MultiNLI \\
   &  & \multicolumn{4}{c}{Performance (Accuracy \%)} \\
  \cmidrule(r){3-6}
  Model & Pretraining & Fixed-rep & Zero-shot & Few-shot{\tiny$(10^3)$} & Many-shot{\tiny \medmuskip=0 mu $(5\cdot10^4)$} \\
  \midrule
  BERT-base      & Unsup.        &  42.4  & 33.3  & 50.9  & 76.4 \\
  BERT-base      & Unsup. + SNLI &  68.0  & 65.9  & 69.6  & 78.2 \\
  BERT-large     & Unsup.        &  38.5  & 33.3  & 61.5  & 79.8 \\
  XLNet-large    & Unsup.        &  48.0  & 33.3  & 76.9  & 84.7 \\
  RoBERTa-large  & Unsup.        &  47.0  & 33.3  & 69.2  & 86.0 \\  
  RoBERTa-large  & Unsup. + SNLI &  81.4  & 78.7  & 81.6  & 85.8 \\  
  \bottomrule
  \end{tabular}
 \end{table}

\section{Continual Learning}
\subsection{Task Settings}
\textbf{Image classification} we split the Tiny-ImageNet \footnote{https://tiny-imagenet.herokuapp.com} dataset into four subsets, each containing examples for 50 classes (500 examples for each class), as tasks 0-3. ``Future'' tasks is chosen to be the original Tiny-ImageNet (200-class joint classification).

\textbf{Language modeling} we first trained an LDA topic model \cite{blei2003latent} on the 1-billion-word-benchmark corpus, and identified 4 topics out of 20: \textit{politics}, \textit{economy}, \textit{medicine}, and \textit{movie}. For each topic we extracted 100,000 sentences as corpus (11MB of text) for the corresponding task. ``Future'' tasks is chosen to be the original 1-billion-word-benchmark corpus.

To examine the generalizable knowledge in models after continual learning, the future task is chosen to be a more general and more complex task, that not only include knowledge in subtasks but also possess a significant amount of new information.

\subsection{Models and Methods}
For our continual learning experiments, we use the ``separate'' design (Table \ref{tab:paradigms}) for dealing with final layers. For L2 and EWC, we optimized a single hyper-parameter: the regularization coefficient $c$, based on average performance. For IMM (weight-transfer) and IMM (l2-transfer), the weighting vector of tasks $\bm{\alpha}$ is chosen to be a uniform weighting, as preliminary optimization of $\bm{\alpha}$ did not yield improved results. The difficulty of optimizing $\bm{\alpha}$ is also discussed in \cite{pfulb2019comprehensive}.

Single-task and multi-task training are used as baselines for comparison. $L_{IA}$ in Table \ref{tab:cont} are all relative to an empty model on the corresponding tasks. 
The performance on the ``future'' task is measured at 10000 training examples.


\subsection{Additional Experiments and Discussion}

\textbf{Another way to inspect forgetting.} In general continual learning settings, the model have no access to past datasets. However, if we are interested in examining the forgetting of representations, we can re-train only the final layer of the network on past tasks to see how it performs. The representation and features of the network (all layers except the final layer) are kept frozen to the value after continual learning. The performance on Tiny-ImageNet task is reported in Table \ref{tab:cont_additional}.

\begin{table}[h]
\setlength{\abovecaptionskip}{7pt}
  \fontsize{8}{9}\selectfont
  \caption{Re-learn the final layers in continual learning. \textit{Acc.(ft)} stands for model accuracy after re-learning the final layer.}
  \label{tab:cont_additional}
  \centering
  \begin{tabular}{lrrrrrrrrrr}
    \toprule
    Tiny-ImageNet \\
     & \multicolumn{2}{c}{Task 0} & \multicolumn{2}{c}{Task 1} & \multicolumn{2}{c}{Task 2} & \multicolumn{2}{c}{Task 3} & \multicolumn{2}{c}{All past}\\
     \cmidrule(r){2-3} \cmidrule(r){4-5} \cmidrule(r){6-7} \cmidrule(r){8-9} \cmidrule(r){10-11}
    Method     & Acc. & Acc.(ft)    & Acc. & Acc.(ft)   & Acc. & Acc.(ft)    & Acc. & Acc.(ft)    & Acc. & Acc.(ft)  \\
    \midrule
       plain   &   7.1   &    53.8 & 17.8  &     53.6& 25.5  &   55.9& 61.1  &  63.2 & 27.9  & 56.7 \\
    L2         &   44.2  &    58.4 & 38.2  &     52.7& 37.8  &   50.2& 45.6  &  50.6 & 41.5  & 53.0  \\
    EWC        &   47.9  &    60.4 & 43.5  &     53.6& 41.6  &   51.6& 44.8  &  50.3 & 44.5  & 54.0 \\
 IMM-mean (wt) &   27.9  &    59.5 & 45.0  &     61.7& 41.5  &   60.3& 41.2  &  54.4 & 38.9  & 59.0 \\
 IMM-mode (wt) &   12.1  &    59.1 & 25.2  &     61.5& 27.3  &   59.8& 21.7  &  53.1 & 21.6  & 58.4 \\
 IMM-mean (l2) &   57.7  &    62.4 & 50.6  &     55.2& 49.1  &   50.6& 48.7  &  50.4 & 51.5  &  54.7 \\
 IMM-mode (l2) &   57.4  &    62.9 & 52.2  &     55.4& 49.6  &   50.8& 47.9  &  49.9 & 51.8  &  54.7  \\
  \midrule
   Single-task &   \multicolumn{2}{c}{56.6}   & \multicolumn{2}{c}{60.7}     & \multicolumn{2}{c}{56.3}     & \multicolumn{2}{c}{53.9}   & \multicolumn{2}{c}{56.9}    \\
    Multi-task &   \multicolumn{2}{c}{62.8}   & \multicolumn{2}{c}{63.6}     & \multicolumn{2}{c}{61.1}     & \multicolumn{2}{c}{61.1}   & \multicolumn{2}{c}{62.2}    \\
    \bottomrule
  \end{tabular}
\end{table}

It is obvious that by only finetuning the final layer (which accounts for less than 2\% of the total number of parameters in the network), the performance can be largely recovered. This indicates that the majority of the information is not forgotten by the network, just as discussed in Section \ref{sec:cont}.

\textbf{Why shuffled-MNIST is not a good task for evaluating forgetting.} Shuffled-MNIST is a commonly used task in evaluating continual learning alglrithms \cite{kirkpatrick2017overcoming,lee2017overcoming,conf/iclr/NguyenLBT18}. In recent study \cite{pfulb2019comprehensive}, it is found that permutation based tasks (like shuffled-MNIST) fail to exhibit forgetting in models and therefore should not be used to evaluate forgetting. From an informational perspective, we can see why: Shuffled-MNIST introduce little new information compared to MNIST (Section \ref{sec:mnist}). Because shuffled-MNIST is just MNIST with labels permuted, the majority of information is shared between the two tasks. The model simply does not have much to forget in transferring from MNIST to shuffled-MNIST, so all models exhibit little forgetting as in \cite{pfulb2019comprehensive}.

\begin{figure}[h]
  \centering
  \includegraphics[width=0.7\linewidth]{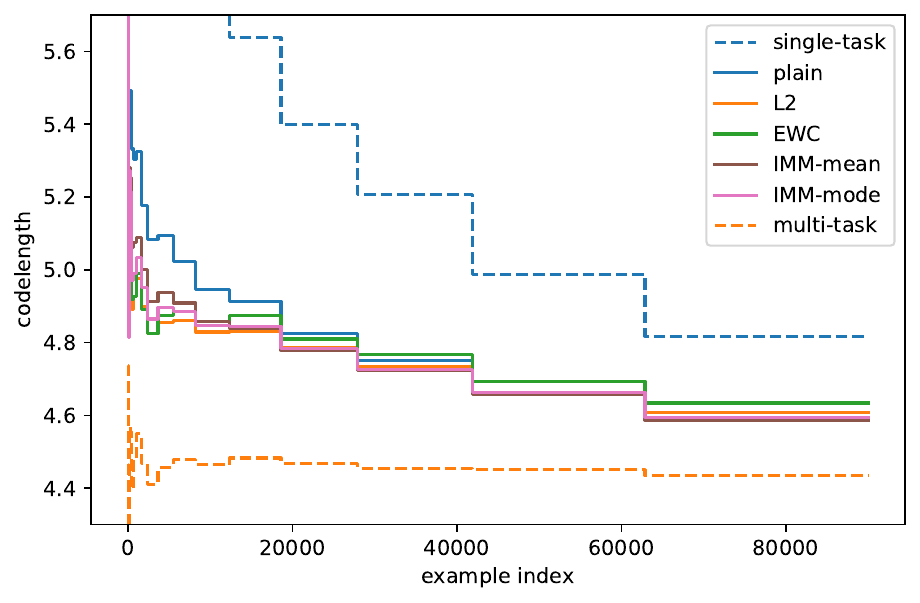}
  \caption{Coding curve on task 0}
  \label{fig:app_orig_cont_0}
\end{figure}

\begin{figure}[h]
  \centering
  \includegraphics[width=0.7\linewidth]{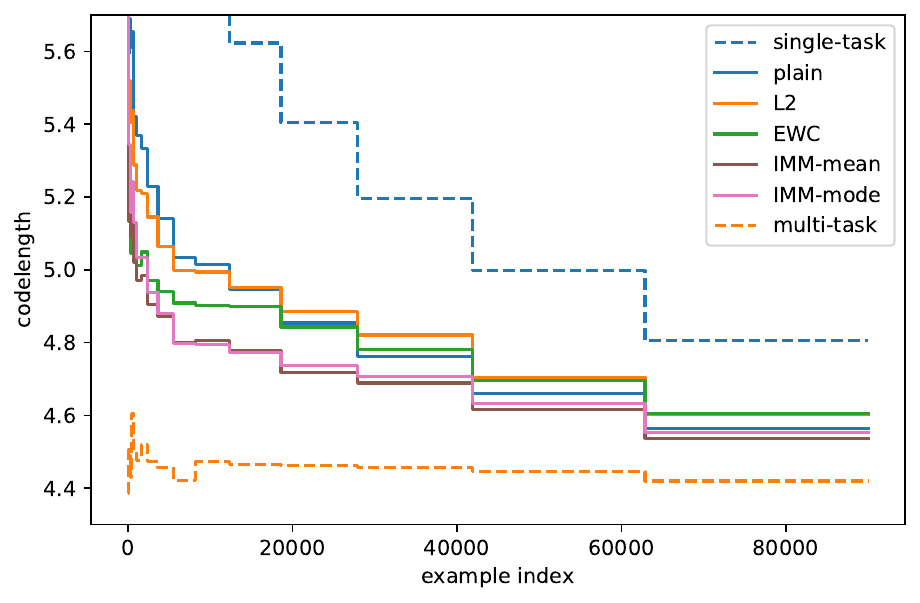}
  \caption{Coding curve on task 1}
  \label{fig:app_orig_cont_1}
\end{figure}

\begin{figure}[h]
  \centering
  \includegraphics[width=0.7\linewidth]{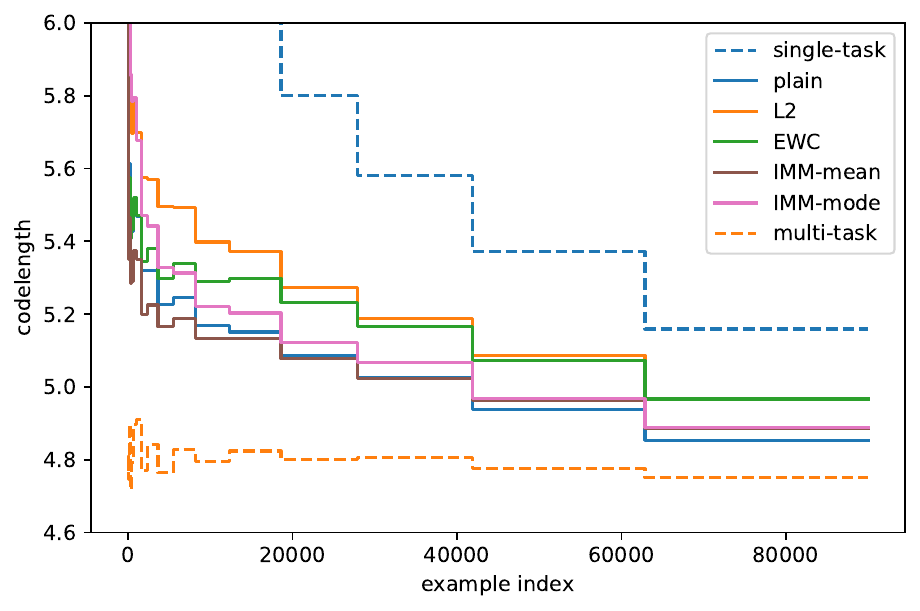}
  \caption{Coding curve on task 2}
  \label{fig:app_orig_cont_2}
\end{figure}

\begin{figure}[h]
  \centering
  \includegraphics[width=0.7\linewidth]{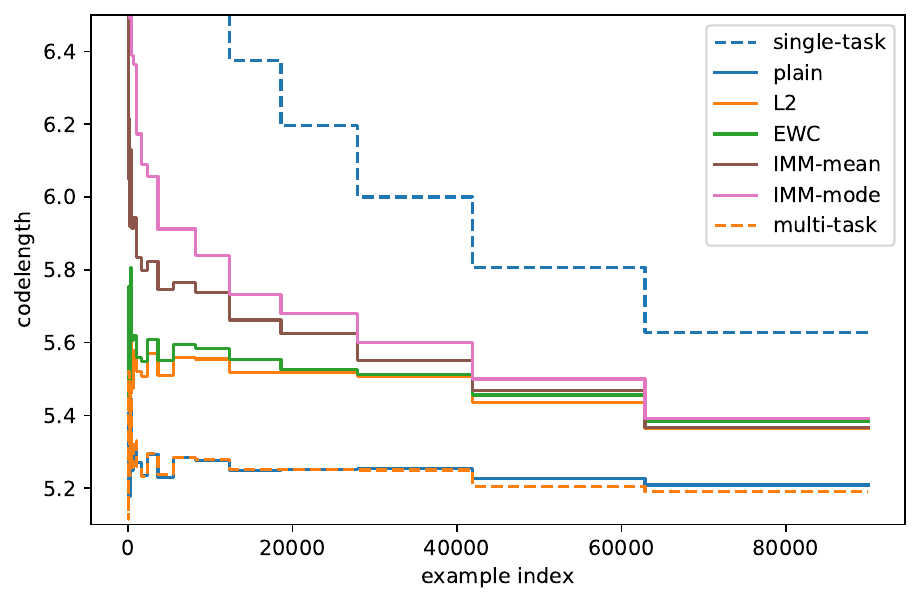}
  \caption{Coding curve on task 3}
  \label{fig:app_orig_cont_3}
\end{figure}

\begin{figure}[h]
  \centering
  \includegraphics[width=0.7\linewidth]{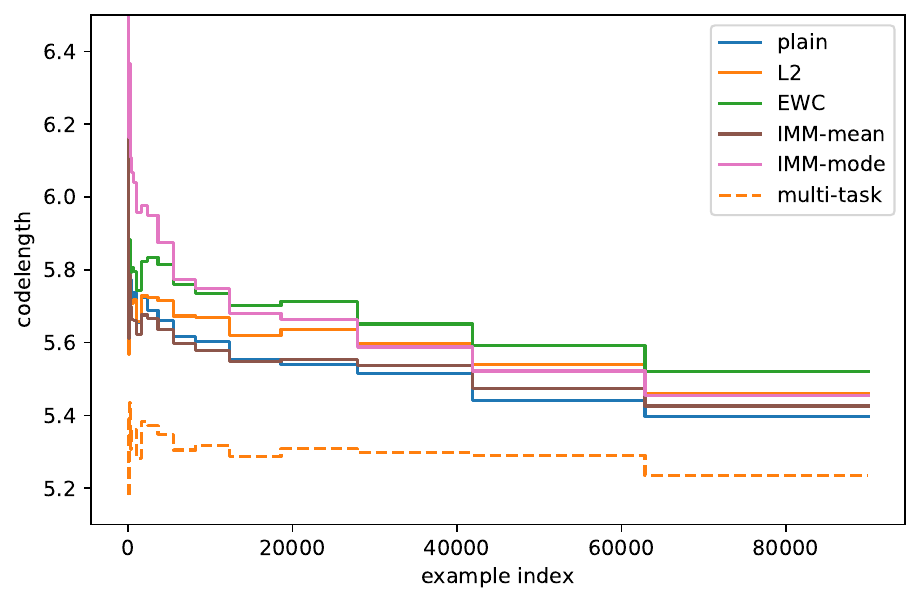}
  \caption{Coding curve on task ``future''}
  \label{fig:app_orig_cont_f}
\end{figure}

\textbf{Raw coding curves.} For language modeling on 1-billion-word-benchmark, we plot the prequential coding curve on each task in Figure \ref{fig:app_orig_cont_0}-\ref{fig:app_orig_cont_f}.

\end{document}